\RequirePackage{snapshot}
\documentclass[10pt,twocolumn,letterpaper]{article}

\usepackage{iccv}
\usepackage{times}
\usepackage{epsfig}
\usepackage{graphicx}
\usepackage{amsmath}
\usepackage{amssymb}

\usepackage{booktabs}
\usepackage[font=small]{caption}
\usepackage{subcaption}
\usepackage{tablefootnote}
\usepackage{adjustbox}
\usepackage{textcomp}  
\usepackage{xcolor,colortbl}  
\usepackage[para]{footmisc}  
\usepackage[square,numbers,sort&compress]{natbib}
\usepackage{tabularx}

\newcommand{\ignore}[1]{}

\usepackage{color}

\newcommand{\tableSize}[0]{\footnotesize}

\renewcommand{\footnotesize}{\scriptsize}

\definecolor{Gray}{gray}{0.9}
\newcolumntype{g}{>{\columncolor{Gray}}c}  
\newcolumntype{H}{>{\setbox0=\hbox\bgroup}c<{\egroup}@{}}  
\definecolor{GrayLine}{gray}{0.7}
\newcommand{\METHOD}[0]{DistInit}
\newcolumntype{Y}{>{\centering\arraybackslash}X}

\usepackage[pagebackref=true,breaklinks=true,letterpaper=true,colorlinks,bookmarks=false]{hyperref}

\iccvfinalcopy 

\def\iccvPaperID{1823}

\ificcvfinal\fi
\begin{document}

\title{\METHOD{}: Learning Video Representations Without a Single Labeled Video}

\author{
Rohit Girdhar$^{1\thanks{Work done as a part of an internship at Facebook}}$ \quad
Du Tran$^{2}$ \quad
Lorenzo Torresani$^{2,3}$ \quad
Deva Ramanan$^{1,4}$ \\
$^{1}$Carnegie Mellon University \quad $^{2}$Facebook \quad $^{3}$Dartmouth College \quad $^{4}$Argo AI \\
}

\maketitle
\thispagestyle{empty}

\begin{abstract}

Video recognition models have progressed significantly over the past few years,
evolving from shallow classifiers trained on hand-crafted features to deep spatiotemporal
networks. 
However, labeled video data
required to train such models have not been able to keep up with the ever-increasing
depth and sophistication of these networks.
In this work, we propose an alternative approach to learning video representations that
require no semantically labeled videos and instead leverages the years of effort
in collecting and labeling large and clean still-image datasets.
We do so by using state-of-the-art models pre-trained on image datasets as ``teachers'' to train video
models in a distillation framework.
We demonstrate that our method learns truly spatiotemporal features, despite being trained only using supervision from
still-image networks. Moreover, it learns good representations across different
input modalities, using completely uncurated raw video data sources and with different
2D teacher models.
Our method obtains strong transfer performance, outperforming standard
techniques for bootstrapping video architectures with image-based models by 16\%. 
We believe that our approach opens up new approaches for learning spatiotemporal representations from unlabeled video data.
\end{abstract}
 \section{Introduction}\label{sec:intro}

\begin{figure}[t]
	\centering
    \includegraphics[width=\linewidth]{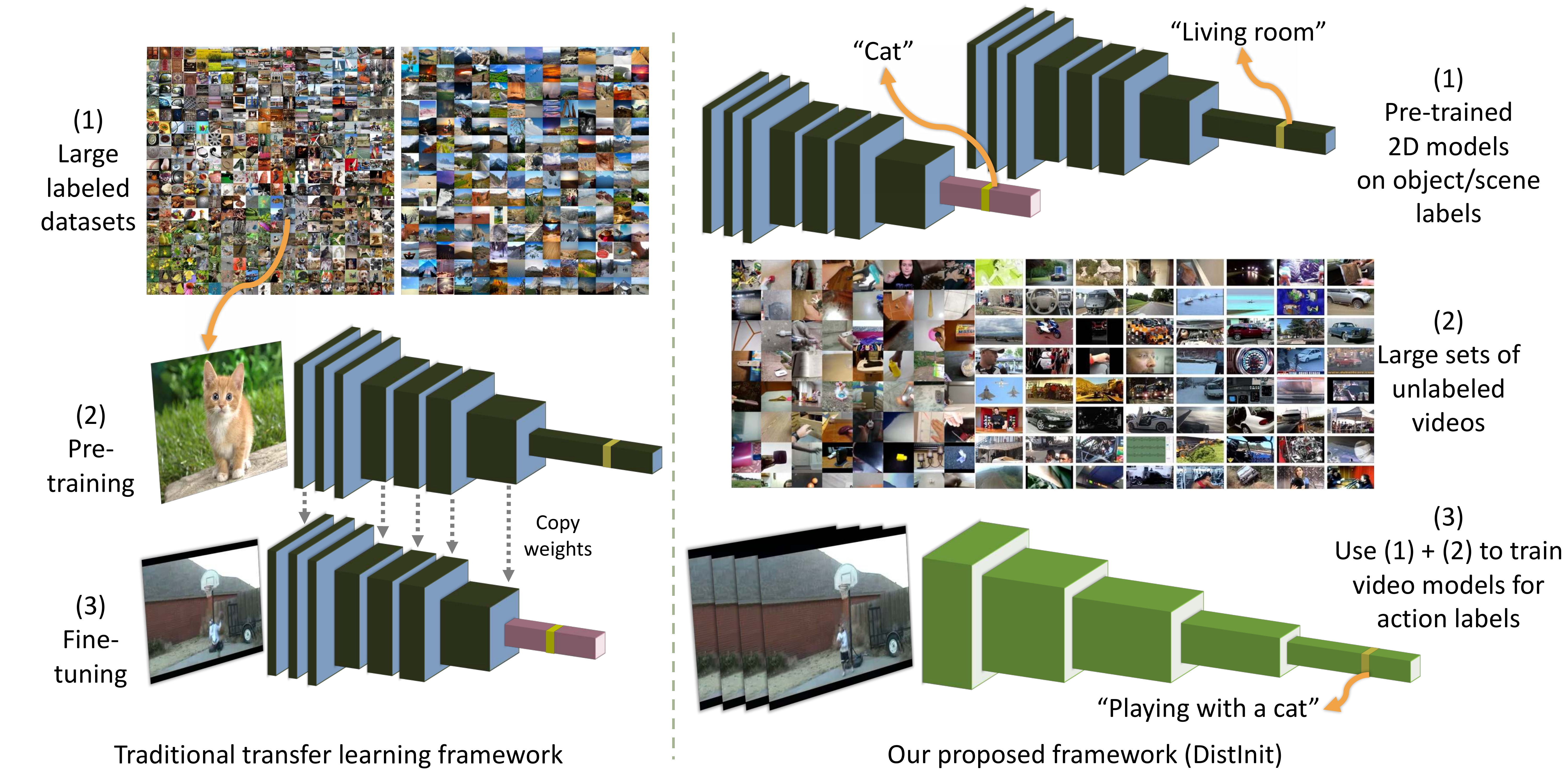}
    \caption{
    {\bf Learning video representations through transfer.}
    Traditional approaches to transfer learning follow the process on left:
    train deep models on large well labeled datasets and finetune on specific task or dataset of choice.
    This approach, while hugely popular, significantly limits the types of models we can use for our specific task, as they must be
    ``compatible'' with the model pre-trained on the large dataset for the learned weights to transfer. This problem is further accentuated in the case of videos, where datasets tend to be small
    or weakly labeled, and models tend to involve 3D/(2+1)D convolutions, making them ``incompatible'' with image models.
    We propose an approach, {\bf \METHOD{}}, to transfer image models to video as shown on the right.
    \METHOD{} starts from models pre-trained on well labeled image datasets with {\em object} or {\em scene} labels, and use them as ``teachers'' for supervising video models. Hence, the video model 
    is able to learn
    spatio-temporal features for video understanding, without needing an explicit {\em action} label for that video.
    }
    \label{fig:teaser}
\end{figure}

Visual recognition tasks have been revolutionized 
by the resurgence of convolutional neural networks (CNNs)~\cite{lecun1998gradient, lecun2015deep} along with the availability of large and well-labeled datasets~\cite{ImageNet,zhou2017places,lin2014microsoft}. This has caused a paradigm shift in computer vision: 
rather than hand-designing better features~\cite{Dalal05,lowe2004distinctive,laptev2005space}, most approaches now train deep models that learn features themselves.
However, deep learning has been transformative not just because models perform well, but because models also {\em transfer}. The dominant illustration of this is the use of ImageNet pre-training~\cite{ImageNet}. It is a near-ubiquitous practice that yields strong improvements across a wide range of tasks, from image classification on small datasets~\cite{krahenbuhl2015data} to pixel-labeling tasks like detection and segmentation~\cite{he2017mask}. Such pre-training is an empirically effective approach to knowledge transfer, where ``knowledge'' is manifested as labeled and curated datasets. 

However, feature learning has not been quite as transformative for video understanding. Early attempts for human action recognition~\cite{Karpathy_14} achieved only marginal improvements over previous state-of-the-art hand-crafted features. Since then, numerous deep architectures~\cite{Simonyan_14b,tran2018closer,Girdhar_17a_ActionVLAD,girdhar2019video,LRCN,Feichtenhofer_16b} have been proposed. 
Interestingly, most performance gains seem to arise from the recent introduction of large-scale video datasets that are carefully curated and annotated~\cite{carreira2017quo,kay2017kinetics}, enabling effective pre-training. Our work introduces simple but novel approaches for pre-training with {\em  un}labeled, {\em un}curated videos. Our approach is motivated by the following two questions:

{\noindent \em 1. What are the ``right'' labels for a video?} Previous work tends to manually defines action ontologies in a top-down fashion
~\cite{ucf101,hmdb51,kay2017kinetics} 
or else discovers classes from bottom-up clustering~\cite{yang2013discovering,fouhey2018vlog}. 
Action classes may be broad -- ``washing" may include both washing one's hands or washing a car~\cite{over2014trecvid}. Classes may also be fine-grained and nuanced, such as ``snuggling with a pillow''~\cite{charades}.
Evidently, the right action vocabulary is unclear and largely up for debate. In contrast, {\em object} labels seem to much better understood, as they can exploit widespread linguistic knowledge bases such as WordNet.
Finally, humans appear able to learn about behaviors and the dynamics of the world even without such explicit action labels.
In this work, we answer this question by {\em making use of objects to label videos}. 

{\noindent \em 2. How do we transfer the knowledge encoded in image datasets~\cite{ImageNet,zhou2017places,lin2014microsoft} into video models?} 
  As discussed earlier, the dominant approach is pre-training. However, because spatiotemporal networks structurally expect a spacetime video as input, they are difficult to (pre)train on images. As a result, many spacetime networks are intentionally designed with an image ``backbone" that allows for pre-training on images. Popular examples include two-stream networks~\cite{Simonyan_14a} and 3D CNNs that ``inflate'' 2D kernels to 3D~\cite{carreira2017quo,Feichtenhofer_17,Feichtenhofer_16b}. However, this places {\em severe} restrictions on the types of video architectures that can be explored. Instead, we introduce a general approach of {\em knowledge transfer by distillation}, which allows us to transfer knowledge from arbitrary image-based teachers to {\em any} spatiotemporal architecture (Fig.~\ref{fig:teaser}). We refer to our approach as {\bf \METHOD{}}.

\METHOD{} leads to a significant 16\% improvement over from-scratch training on the HMDB dataset, getting almost half-way to the improvement provided by pretraining on a fully-supervised dataset like Kinetics~\cite{kay2017kinetics}. From-scratch training is the defacto standard for state-of-the-art architectures that can not be initialized or inflated from image architectures~\cite{tran2018closer}. While large-scale video datasets like Kinetics now provide an alternate path for pre-training, \METHOD{} does so without requiring {\em any} video data curation. As we show in Section~\ref{sec:expt:distill_data}, it is able to learn competitive representations from an internal uncurated dataset of random web videos. This is in contrast to previous works~\cite{noroozi2018boosting,doersch2015unsupervised,gidaris2018unsupervised} on unsupervised learning that use ImageNet without labels but still potentially benefit from the data curation.
 \section{Related Work}\label{sec:relwork}

{\noindent \bf Feature learning:} Video understanding, specifically for the task of human action recognition, is a well studied
problem in computer vision. Analogously to the progress of image-based recognition methods, which have advanced
from hand-crafted features~\cite{lowe2004distinctive,Dalal05} to modern deep networks~\cite{Szegedy_16,He_16,Simonyan_14a},
video understanding methods have also evolved from hand-designed models~\cite{IDT_Wang_13,WangCVPR11,laptev2005space}
to deep spatiotemporal networks~\cite{Tran_15,Simonyan_14b}.
However, while image based recognition has seen dramatic gains in accuracy, improvements in video analysis have been
more modest. 
In the still-image domain, deep models have greatly benefited from the availability of well-labeled datasets, such as ImageNet~\cite{ImageNet} or Places~\cite{zhou2017places}.

{\noindent \bf Video datasets:} Until recently, video datasets have either been well-labeled but small~\cite{hmdb51,ucf101,charades},
or large but weakly-labeled~\cite{Karpathy_14,youtube8M}. A recently introduced dataset, Kinetics~\cite{kay2017kinetics},
is currently the largest well-annotated dataset, with around 300K videos labeled into 400 categories (we note a larger version with 600K videos in 600 categories was recently released).
It is  nearly
two orders of magnitude larger than previously established  benchmarks in video classification~\cite{hmdb51,ucf101}. As expected, pre-training
networks on this dataset has yielded significant gains in accuracy~\cite{carreira2017quo} on many standard benchmarks~\cite{hmdb51,ucf101,charades},
and have won CVPR 2017 ActivityNet and Charades challenges.
However, it is worth noting that this dataset was collected 
at a significant curation and annotation effort~\cite{kay2017kinetics}. 

{\noindent \bf Video labels:} The challenge in generating large-scale well-labeled video datasets stems from the fact that a human annotator has to spend much longer to label a  video compared to a single image. Previous work has attempted to reduce this labeling effort through heuristics~\cite{zhao2017slac}, 
but these methods still require a human annotator to clean up the final labels. There has also been some work in learning unsupervised video
representations~\cite{Sermanet2017TCN,misra2016unsupervised}, however has typically lead to inferior results compared to supervised features.

{\noindent \bf Pre-training:} The question we pose is: since labeling images is faster, and since we already have large, well-labeled image datasets such as ImageNet, can we instead use
these to bootstrap the learning of spatiotemporal video architectures? Unsurprisingly, various previous approaches have attempted this. The popular
two-stream architecture~\cite{Simonyan_14b} uses individual frames from the video as input. Hence it initializes the 
RGB stream of the network with
 weights pre-trained on ImageNet and then fine-tunes them for action classification on the action dataset. More recent variants of
two-stream architectures have also initialized the flow stream~\cite{WangL_16a} from weights pretrained on ImageNet by viewing  optical flow as a grayscale image.

{\noindent \bf Inflation:} However, such initializations are only applicable to video models that use 2D convolutions, analogous to those applied in CNNs for still-images. 
What about more complex, truly spatiotemporal models, such as 3D convolutional architectures~\cite{Tran_15}?
Until recently, such models have largely been limited to pre-training on large but weakly-labeled video datasets, such as Sports1M~\cite{Karpathy_14}.
Recent work~\cite{carreira2017quo,Feichtenhofer_16b} proposed a nice alternative, consisting of inflating standard 2D CNNs kernels to 3D, by simply replicating the 2D kernels in time. While effective in getting strong performance on large benchmarks, on small datasets this approach tends to bias video models to be close to static replicas of the image models. 
Moreover, such initialization constrains the 3D architecture to be identical to the 2D CNN, except for the additional third dimension in kernels. This effectively restricts the design of video models to 
extensions of what
works best in the still-image domain,
which may not be the architectures for video analysis.

{\noindent \bf Distillation:} 
Our approach is inspired by techniques that distill teacher networks into student models~\cite{hinton2015distilling}. 
However, distillation typically trains smaller student models on the same data distribution used to train the teacher (with the goal of increasing efficiency). Our approach instead focuses on representation learning through pre-training. Our students are larger (3D vs 2D CNNs) and geared for different data domains (videos vs images).
The most related work may be cross-modal distillation~\cite{gupta2016cross}, which transfers supervision from RGB to flow or depth modalities. Importantly, such work focuses on the {\em same} end task, such as object detection. In contrast, we focus on {\em task distillation}, where tasks are quite different (object detection versus action classification). 
From this perspective, our philosophy aligns with taskonomies~\cite{zamir2018taskonomy}, which advocates the use of different pre-training tasks for a variety of target tasks. But rather than advocating pre-training, we pursue distillation since it 
allows for arbitrary changes in network topology.
Other works have also used similar distillation frameworks for action recognition tasks, such as~\cite{diba2018spatio} which transfers supervision from frames to videos by solving a correspondence problem. Our approach is much simpler, and directly transfers supervision by matching label distributions. Finally, prior works have also shown improvements using the 2D image-based networks (`teachers', in our context) directly, as additional features for action recognition~\cite{diba2017deep,girdhar2018better}, hence reinforcing our observation that scene/object information is a very useful cue
for video understanding.

{\noindent \bf Domain adaptation:} Our work is also related to techniques for domain adaptation (DA)~\cite{pan2009domain,saenko2010adapting}, where the goal is adapting a network to a new data distribution. Our formulation differs in that we also adapt to new tasks (object classification vs action recognition) and network architectures (2D CNNs vs 3D CNNs).
We show extensive
experiments with standard benchmarks and show significant improvements over inflation and other previous approaches
in learning video representations for action recognition. \section{Our Approach}\label{sec:approach}

\begin{figure}[t]
	\centering
    \includegraphics[width=\linewidth]{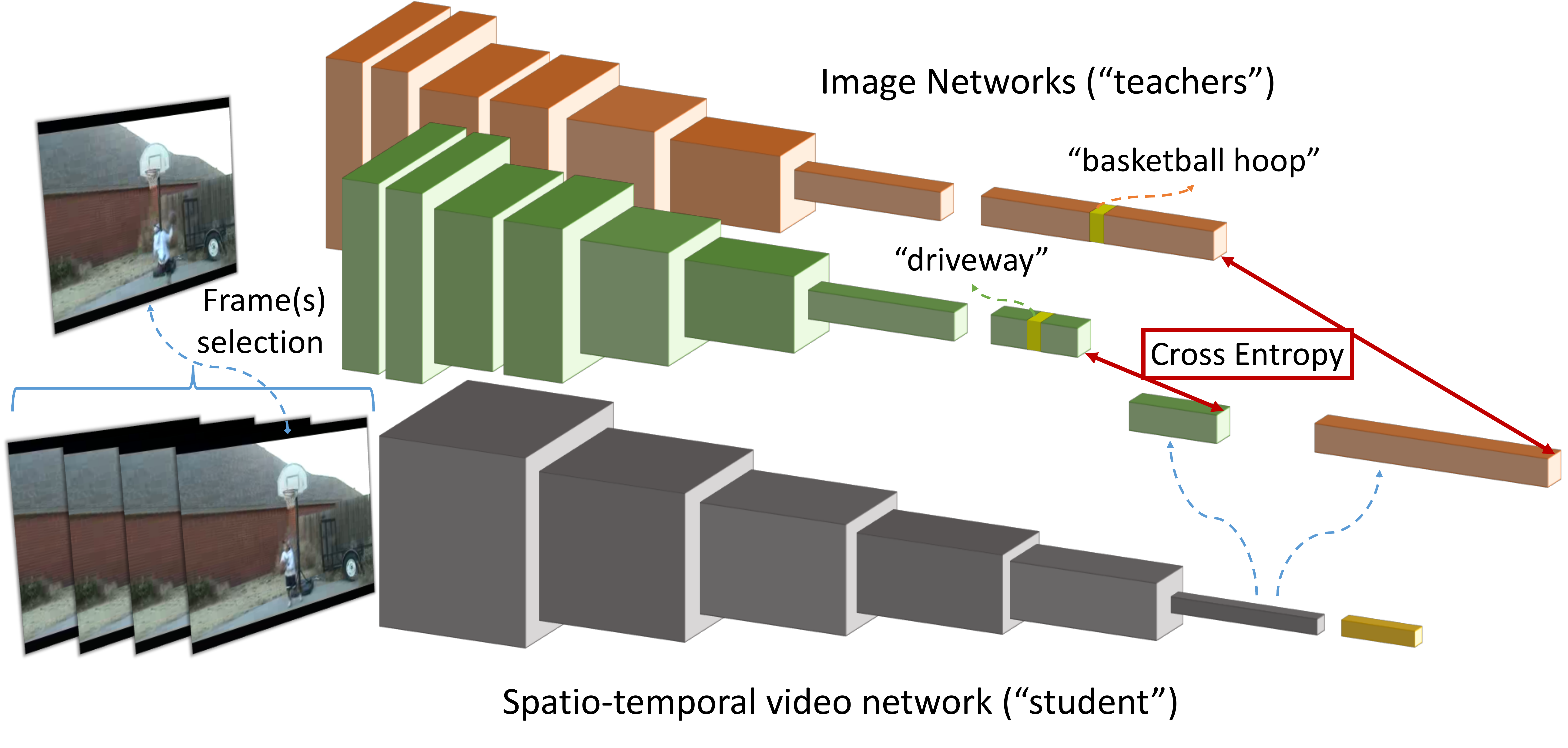}
    \caption{{\bf \METHOD{} network architecture.} We use random frames from the input clip to generate soft-labels for the video model, using an arbitrary number of image-based teachers networks. The student tries to match the targets provided by the  teachers.}
    \label{fig:nwarch}
\end{figure}

We now describe our approach in detail. To reiterate, our goal is to learn video representations without using any video annotations.
We do so by leveraging pre-trained 2D networks, using them to supervise or ``teach'' the video models.
Hence, we refer to the 2D pre-trained networks as ``teachers'' and our target video network as ``student''. We make no assumption over the respective architectures of these models, i.e., we do no constrain the structure of the 3D network to be merely a 3D version of the 2D networks it learns from or to have a structure compatible with them. 

Figure~\ref{fig:nwarch} depicts the network architecture used to train the student network. We start with teacher networks trained on standard image-level tasks, such as image classification on ImageNet. While in this work we primarily focus on classification, our architecture is generic and can also benefit from teachers trained on spatial tasks such as detection, keypoint estimation and so on,
with the only difference being the definition of the distillation loss function.
Also, our architecture is naturally amenable to work with an arbitrary number of teachers, which can be used in a multi-task learning framework to distill information from multiple domains into the student.  Throughout the training process, these teacher networks are kept fixed, in ``test'' mode,
and are used to extract a feature representation from the video to be used as a ``target''
to supervise the student network.

Since teacher networks are designed to find {\em objects
} in images, it is not obvious how to use them to extract features for {\em actions} in video. We propose a 
simple 
solution: pre-train the spatiotemporal action network for finding objects in frames of a video.
However, our teacher networks are designed to work over images, 
so how do we apply them on a video?
We experiment with standard
approaches from the literature, including uniform or random sampling of frames~\cite{Simonyan_14b},
as well as averaging predictions from multiple different frames~\cite{WangL_16a}.
In this work we use the last-layer features in the form of normalized softmax predictions or (unnormalized) logits.   
In case of multiple
frames, we average the teacher logits before computing a normalized prediction target.
The {\em student} network then takes the complete video clip as input.
We train it to be able to predict the features or probability distribution produced by the teacher.
For this purpose, we define the last layer in the student network to be a linear layer that takes the final spatiotemporally averaged feature tensor and maps it to a number of units that matches the dimensionality of the output generated by the teacher. In case of multiple teachers, we define 
a linear layer per teacher, and
optimize all losses jointly.

To formalize, let us denote a video as $x = \{x_1,\ldots x_T\}$, where $x_t$ is the $t^{th}$ frame. In our problem formulation, we have access to a teacher that reports a prediction label at each frame. For simplicity, we assume that the teacher returns a (softmax) distribution over $K$ classification labels. We generate a distribution over labels for a video by (1) averaging the per-frame logits for the $k^{th}$ class $z_k(x_t)$ and (2) passing the average through a softmax function with temprature $\tau$ (typically $\tau=1$):
\begin{align}
y_k(x) =  \frac{e^{\frac{1}{\tau T}\sum_t z_k(x_t)}}{\sum_l e^{\frac{1}{\tau T}\sum_t z_l(x_t)}} \quad \text{\bf [Temporal Averaging]} 
\end{align}
The resulting distribution is then used as soft targets for training weights $w$ associated with a student network $f$ of arbitrary architecture by means of the following objective:
\begin{align}
     \text{Loss}(w) = -\sum_k y_k(x) \log f_k(x;w) \quad \text{\bf [Soft Targets]} 
\end{align}
\noindent where $f_k(x;w)$ is the student softmax distribution for label $k$. Finally, we explore multi-source knowledge distillation by adding together losses from different image-based teachers:
\begin{align}
     \min_w \sum_i\text{Loss}_i(w) \quad \text{\bf [Multi-Source Distillation]}
\end{align}
In our experiments, we explore teachers trained for object classification (ImageNet) and scene classification (Places).

We train the network using loss functions inspired from the network distillation literature~\cite{bucilua2006model,hinton2015distilling}. When using the teacher network to produce a probability
distribution, we train the student to produce a matching distribution by incurring a cross entropy loss between the two distributions.
As suggested in~\cite{hinton2015distilling}, we also tried using different
values of {\em temperature} ($\tau$) to scale the logits before computing softmax and cross entropy, but found temperature
value of 1 to perform best in our experiments. We also experimented with the a mean squared loss on the logits (before
softmax normalization), as suggested in~\cite{bucilua2006model}, and compare performance in Section~\ref{sec:expt}.

{\bf \noindent Architecture Details:}
We use recent, state-of-the-art, network architectures for all experiments and comparisons. For the still-image {\em teacher} networks, we use the  ResNet-50~\cite{He_16} architecture, trained on different image datasets such ImageNet 1K~\cite{ImageNet} and Places 365~\cite{zhou2017places}. For the spatiotemporal (video) {\em student} architectures, we first experiment with a variant of the Res3D~\cite{tran2017convnet} architecture. Res3D is an improved version of the popular
C3D~\cite{Tran_15} using residual connections. We denote a $N$-layer Res3D model as Res3D-$N$, which is compatible with the standard ResNet-$N$~\cite{He_16} architecture. Since there is a one-to-one correspondence between such 2D and 3D models,
the 3D models can also be initialized by {\em inflating} the learned weights from 2D models (e.g., for each channel, replicate the 2D filter weights along the temporal dimension to produce a 3D convolutional filter). Similar ideas of inflating 2D models to 3D have  been proposed previously for Inception-style architectures~\cite{carreira2017quo}, along with initialization techniques from corresponding 2D models~\cite{carreira2017quo,Feichtenhofer_17,Feichtenhofer_16b}. The existence of a 1-to-1 mapping between the 2D and 3D models used in our experiments allows us to compare our approach to the method of inflation for initialization. However, we stress that unlike inflation, our method is applicable even in scenarios where such 1-to-1 mapping does not hold.

{\noindent \bf (2+1)D CNNs:} More recently, full 3D models have been superseded by (2+1)D architectures~\cite{tran2018closer},
where each 3D kernel is decomposed into a 2D spatial component followed by a 1D temporal filter. 
Similar models have also been proposed previously~\cite{sun2015human}, and are also known as P3D~\cite{qiu2017learning} or S3D~\cite{xie2017rethinking} architectures.
These models have proven to be more efficient, with much fewer parameters, and more effective on various standard benchmarks~\cite{xie2017rethinking,tran2018closer}.
However, these models no longer conform to standard 2D architectures because they contain additional \texttt{conv} and \texttt{batch\_norm} layers that extend over time. These parameters do not exist in
corresponding 2D models and so cannot be initialized with images. Nevertheless, our distillation remains applicable even in this scenario. In this work, we refer to such networks using R(2+1)D-$N$ notation, for $N$-deep architecture.

{\bf \noindent Implementation Details:}
For all experiments, we use the hyperparameter values described in~\cite{tran2018closer}.
For distillation pre-training, we use the hyper-parameter setup for ``Kinetics from-scratch training.'' We use distributed Sync-SGD~\cite{goyal2017accurate} over $16\times 4$ GPUs, starting
with LR=0.01, and dropping it by 10$\times$ every 10 epochs. Weight decay is set to $10^{-4}$.
We train for a total of 45 epochs, where each epoch is defined as 1M iterations.
The video model is learned on 8-frames clips of 112 pixels. The network has
depth of 18, which enables faster experimentation compared  to the best model reported in~\cite{tran2018closer} which uses 32 frames and has a depth of 34 layers.
The batch size used for Kinetics training is 32/GPU, which we reduce to 24/GPU to accommodate the additional
memory requirements for the teacher networks.
For the finetuning experiment on smaller datasets like HMDB, we use Sync-SGD with $8\times 2$ GPUs, starting
with LR=0.002, an dropping it by $10\times$ every 2 epochs. The weight decay is set to $5\times 10^{-3}$. We train 8 epochs, with each epoch defined as 40K steps. 
When training from scratch, we use initial LR of 0.01 with a step every 10 epochs, trained for total of 45 epochs.
 \section{Experiments}\label{sec:expt}

We now experimentally evaluate our system. We start by introducing the datasets and benchmarks used for training and evaluation in Section~\ref{sec:expt:data}. We then compare \METHOD{} with inflating 2D models for initialization in Section~\ref{sec:expt:inflate}. Next we ablate the various design choices in \METHOD{} in Section~\ref{sec:expt:diagnostics}, and finally compare to previous state of the art on UCF-101~\cite{ucf101} and HMDB-51~\cite{hmdb51} in Section~\ref{sec:expt:sota}.

\subsection{Datasets and Evaluation}\label{sec:expt:data}

Our method involves two stages, as typical in video understanding literature~\cite{carreira2017quo}: pre-training on a large, unlabeled corpus of videos using still-image models as ``teachers'', followed by fine-tuning on the training split of a labeled target dataset (`test bed'). After training, we evaluate the performance
on the test set of the target dataset. 

{\noindent \bf Unlabeled source videos:} We experiment with a variety of different unlabeled video corpuses in Section~\ref{sec:expt:distill_data}, including Kinetics~\cite{kay2017kinetics}, Sports1M~\cite{Karpathy_14} 
and an internal set of videos.
While some of these datasets do come with semantic (action) labels, we ignore such annotations and only use the raw videos.
Kinetics
and Sports1M contain about 300K
and 1.1M videos, respectively.
In this work, we drop any labels these datasets come with, and only use the videos as a large, unlabeled corpus to train video representations.
The internal video set includes 1M videos without any semantic labels and randomly sampled from a larger collection. We use these diverse datasets to show that our method is not limited to any form of data curation, and can work with truly in-the-wild videos.

{\noindent \bf Target test videos:} HMDB-51~\cite{hmdb51} contains 6766 realistic and varied video clips from 51 action classes. Evaluation is
performed using average classification accuracy over three train/test splits from~\cite{THUMOS13}, each with 3570
train and 1530 test videos. UCF-101~\cite{ucf101} consists of 13320 sports video clips from 101 action classes,
also evaluated using average classification accuracy over 3 splits. We use the HMDB split 1 for ablative experiments, and report the final
performance on all splits for HMDB and UCF in Section~\ref{sec:expt:sota}.

\begin{figure*}[t]
    \centering
    \begin{tabularx}{\linewidth}{YYY@{\hskip 1cm}YYY}
    $t=1$ & $t=2$ & $t=3$ & $t=1$ & $t=2$ & $t=3$
    \end{tabularx}
    \includegraphics[trim={0 2.5cm 0 0},width=\linewidth]{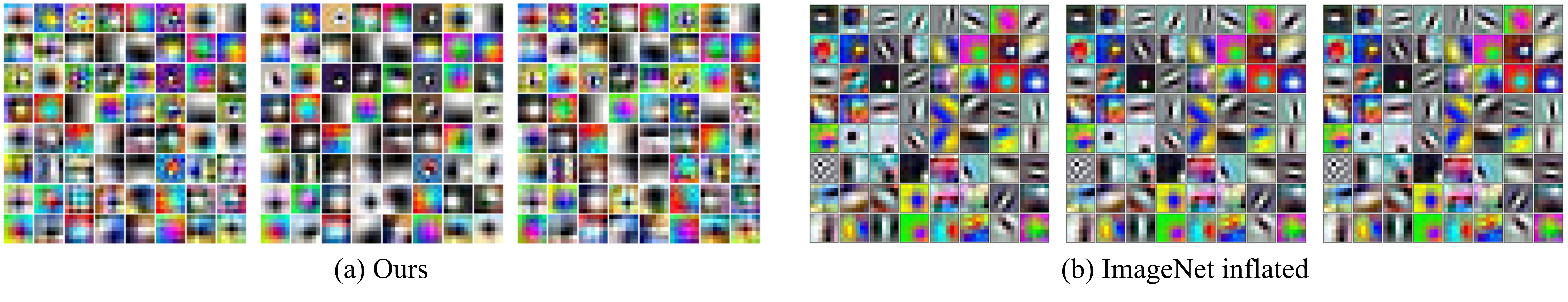}
    \caption{
    {\bf Learned Res3D filters.}
    We compare the learned first layer representation using our distillation approach, to inflation.
    For each, we show the 64 \texttt{conv\_1} filters, for each time instance of the filter.
    As described in Section~\ref{sec:expt:inflate}, our filters change in value over time, indicating that they learn to look for some amount of temporal dynamics in the input video. This clearly contrasts with ImageNet-inflated filters, which are {\em exact} copies over time, and so do not respond to any temporal change in pixel values.
    }
    \label{fig:filters_res3d}
\end{figure*}

\begin{table}[t]
\setlength\tabcolsep{6pt} 
\tableSize{}
\centering
\begin{tabular}{llrgr}
\toprule
Model & Initialization & Per clip & Top 1 & Top 5 \\
\midrule
Res3D-$18$ & Scratch & 24.6 & 25.4 & 55.2 \\  
Res3D-$18$ & ImageNet inflated & 32.5 & 35.8 & 66.2 \\  
Res3D-$18$ & PlaceNet inflated & 32.5 & 35.6 & 66.2 \\  
Res3D-$18$ & DistInit (ours) & {\bf 36.6} & {\bf 39.9} & {\bf 73.5} \\  
\arrayrulecolor{GrayLine}
\midrule
R(2+1)D-$18$ & Scratch & 22.0 & 24.1 & 53.1 \\  
R(2+1)D-$18$ & DistInit (ours) & {\bf 37.8} & {\bf 40.3} & {\bf 74.4} \\  
\midrule
R(2+1)D-$18$ & Kinetics pre-training & - & 51.0 & - \\
\arrayrulecolor{black}
\bottomrule
\end{tabular}
\caption{
{\bf Distillation vs Inflation.}
As described in Section~\ref{sec:expt:inflate}, our distillation approach
outperforms training video models from scratch or initializing 
them by inflating 2D models. We evaluate using
percentage accuracy on the HMDB-51 dataset, Split 1.
The models used are 18-layer Res3D and R(2+1)D, over 8-frame input,
trained with cross-entropy loss (described in Section~\ref{sec:expt:loss}).
The DistInit training is done using 2D network trained on ImageNet.
}\label{tab:inflate}
\end{table}

\subsection{\METHOD{} vs Inflation}\label{sec:expt:inflate}
{\noindent \bf Inflation:} We first compare our proposed approach to inflation~\cite{carreira2017quo,Feichtenhofer_16b}, i.e., initializing video models from 2D models by inflating 2D kernels to 3D via replication over time. Note that inflation is constrained to operate on 3D models that have a one-to-one correspondence with the 2D model. Hence, we use a Res3D-18 model, which is compatible for direct inflation from ResNet-18 models. We experiment with publicly available ImageNet and PlaceNet models. We compare it with our distillation approach in Table~\ref{tab:inflate}, trained using an ImageNet pretrained model as the teacher. Distillation improves performance by 15\% over a model trained from scratch, and 4\% over a model trained with inflated weights (the current best-practice for training such models). 

{\noindent \bf (2+1)D:} More importantly, our approach can also be used to initialize state-of-the-art temporal architectures such as R(2+1)D~\cite{tran2018closer}, which do not have a natural 2D counterpart. In such a setting, the current best practice is to initialize such networks from scratch. Here, distillation improves performance by 16\%. Finally, we also report the model trained using actual Kinetics labels, and as expected, that yields higher performance. Hence there is clear value to the explicit manual supervision provided in such large-scale datasets, but distillation appears to get us ``half-way'' there.

\begin{figure}[t]
    \centering
    \includegraphics[width=\linewidth]{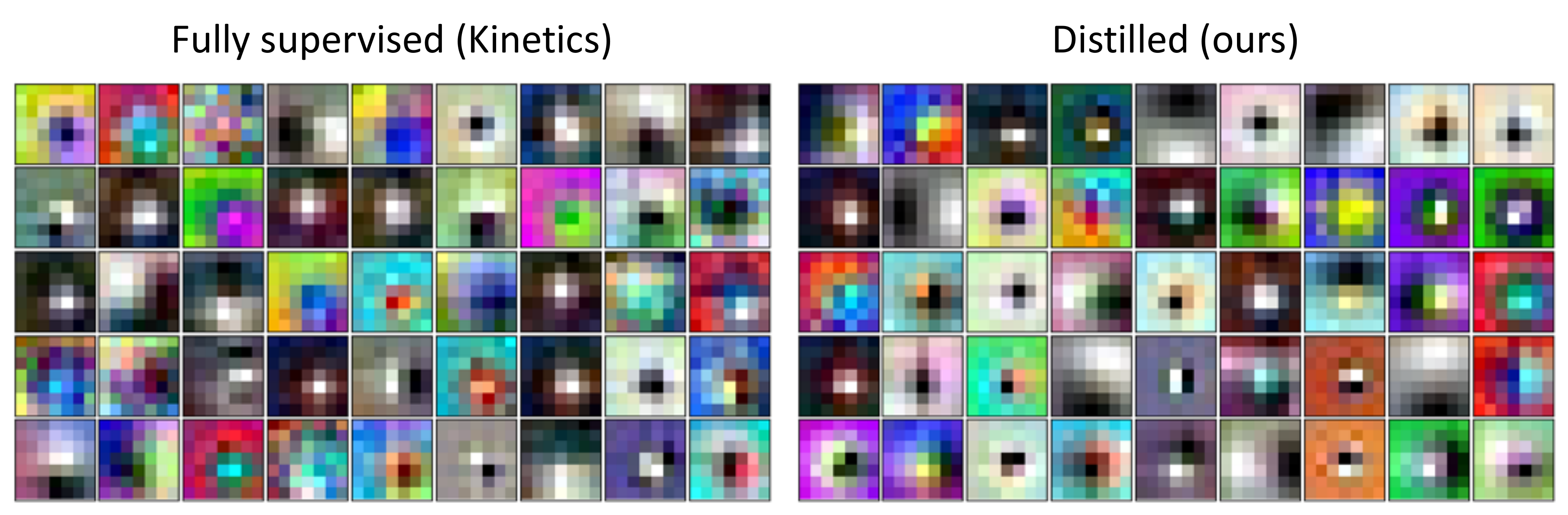}
    \caption{
    {\bf Learned R(2+1)D filters.}
    Similar to Figure~\ref{fig:filters_res3d}, we show the first layer conv filters for the R(2+1)D models.
    Note that 2.5D conv layer contains a 2D convolution in space followed by 1D convolution in time, and in this visualization
    we are only showing the former, i.e. the 45 2D conv filters that operate on the RGB image. We observe that our distillation approach learns spatiotemporal representations that are relatively
    similar to the fully supervised model (compared to 
    filters learned from Imagenet in Fig.~\ref{fig:filters_res3d}).
    }
    \label{fig:filters_R(2+1)D}
\end{figure}

{\noindent \bf Visualizations:} At this point, it is natural to ask why the distilled model outperforms current best-practices such as inflation? We visualize the learned representation by plotting the first layer \texttt{conv} filters in Figure~\ref{fig:filters_res3d}. It can be seen that our distilled model learns truly spatiotemporal filters that vary in time, whereas inflation simply {\em copies} the same filter over time. Such dynamic temporal variation is readily present in the videos used for distillation, even when they are not labelled with spatiotemporal action categories. Filters pre-trained with inflation initialization never see actual video data, and so cannot encode such variation. In Figure~\ref{fig:filters_R(2+1)D} we also compare the filters learned by our R(2+1)D model via distillation vs via fully-supervised training. Our filters look quite similar to those learned through supervised learning, showing the effectiveness of our approach. In some sense, the improved performance of distillation can be readily explained by more data --  networks learning from scratch see no data for pre-training, inflation networks see ImageNet, while distilled networks see both Imagenet and unlabeled videos. Our practical observation is that one can use image-based teachers to pre-train on massively large, unlabeled video datasets.

\begin{figure}
    \centering
    \includegraphics[width=\linewidth]{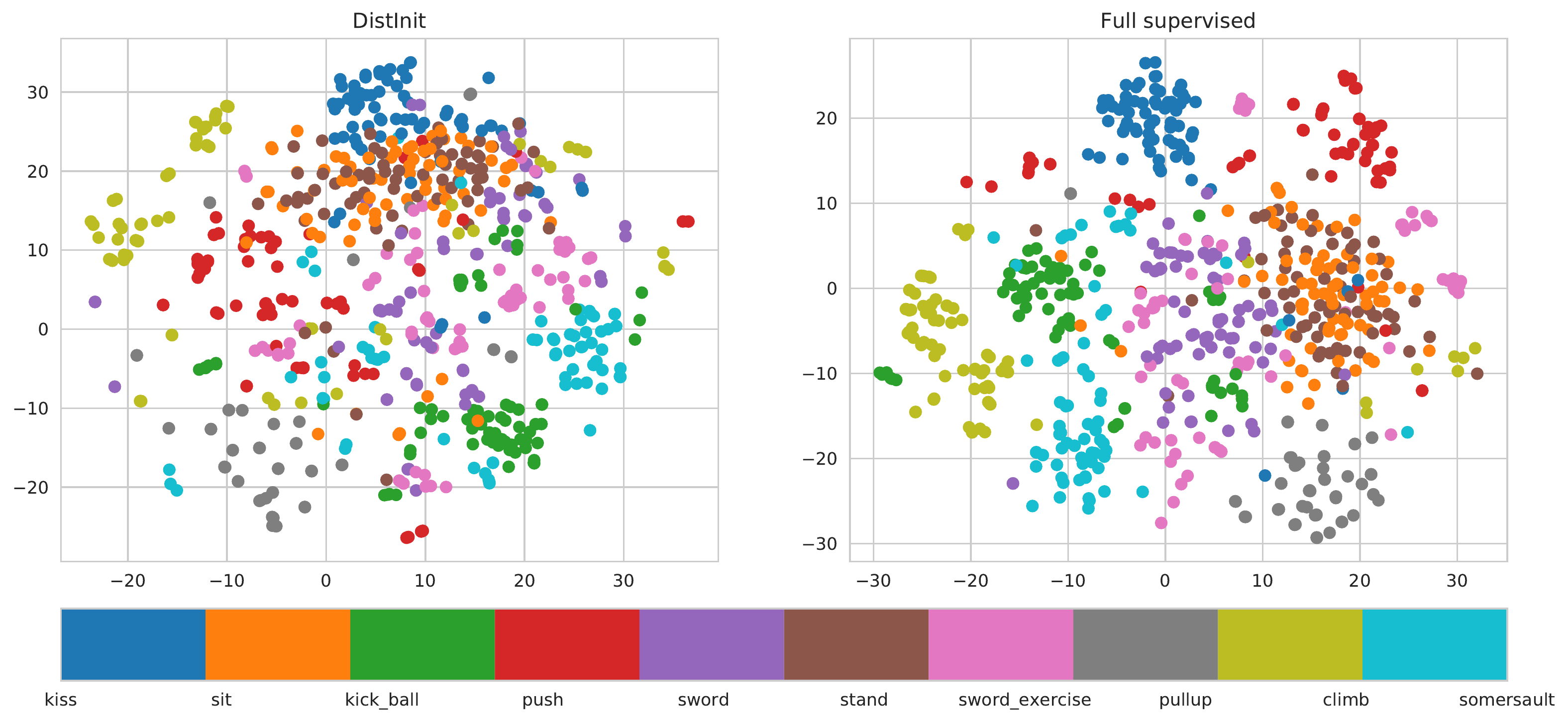}
    \caption{{\bf Learned high-level representation.}
    While the filter maps in  Figure~\ref{fig:filters_res3d} and \ref{fig:filters_R(2+1)D} can be used to interpret
    the low-level representation learned by our model, we now try to probe the high level representation by visualizing the 
    last layer features.
    This figure shows tSNE~\cite{tsne} visualization of averaged last layer features from the model trained with \METHOD{}, and trained
    with full Kinetics supervision. Each dot represents a video from HMDB training set, and is color coded by
    the class of that video. For ease of visualization, we picked 10 {\bf random} classes to plot.
    Note that \METHOD{} is already able to segregate many videos into clusters correlated with their action classes, without ever being trained on any action labels!
    The fully supervised model naturally does better as it has been trained on a large action dataset, Kinetics.
    This further suggests \METHOD{} leads to useful representation for classifying actions.
    }
    \label{fig:tsne}
\end{figure}

We can also analyze the effectiveness of distillation pre-training, by visualizing the correlation of the representation we learn with the classes in the task of action recognition. As explained
in Figure~\ref{fig:tsne}, we can see that the last layer features for the 
same action class tend to cluster together when projected to 2D using tSNE~\cite{tsne}. 

\subsection{Diagnostic Analysis}\label{sec:expt:diagnostics}

{\noindent \bf Design choices for the teacher network:}
We now ablate the design choices for the teacher networks. Teacher networks are required to generate a target label to
supervise the video model being trained, by using images from the video clip. We experiment with picking the center frame, a random frame, or multiple random frames from the clip to compute the targets. In case of multiple frames, we average the logits before
passing them through softmax to generate the target distribution. We compare these methods in Table~\ref{tab:pick_strategy},
and observe higher performance when picking frames randomly. This improvement 
can be due
to less overfitting through label augmentation. We use it in our final model.

\begin{table}[t]
\setlength\tabcolsep{6pt} 
\tableSize{}
\centering
\begin{tabular}{llrgr}
\toprule
Model & Pick strategy & Per clip & Top 1 & Top 5 \\
\midrule
R(2+1)D-$18$ & Center & 37.8 & 40.3 & 74.4 \\  
R(2+1)D-$18$ & Random & 39.9 & 43.2 & 73.9 \\  
R(2+1)D-$18$ & 2 Random & 39.6 & {\bf 44.0} & 73.5 \\  
\bottomrule
\end{tabular}
\caption{
{\bf Video to Image.}
We compare different strategies of converting the video into image(s) for extracting the target label. We find strongest
performance when picking random frames to generate the target distribution.
Model used here is 18-layer R(2+1)D, over 8-frame input,
trained with cross-entropy loss (Section~\ref{sec:expt:loss}); evaluated using percentage accuracy on HMDB-51 split 1.
}\label{tab:pick_strategy}
\end{table}

{\noindent \bf Distillation loss:}\label{sec:expt:loss}
Next, we evaluate the different choices for the loss function in distillation. As already explained
in Section~\ref{sec:approach}, previous work has suggested different loss functions for distillation tasks.
We compare two popular approaches: KL divergence over distribution and $l_{2}$ loss over logits. In the case of the former,
we compute the softmax distribution from the teacher networks, as well as from the student branch that attempts 
to match that teacher, and use a cross entropy between the two softmax distributions as the objective to optimize.
We find this objective can be well optimized using the standard hyper-parameter setup used for Kinetics training
in~\cite{tran2018closer}.
In the case of the latter, we skip the softmax normalization step and directly compute the mean squared error between the last
linear layers as the objective. Since the initial loss values are much higher,
we needed to drop the learning rate by a factor of 10 to optimize this model, with all the other parameters kept the same.
As Table~\ref{tab:loss_fn} shows, we observe nearly similar downstream performance with both.

\begin{table}[t]
\setlength\tabcolsep{6pt} 
\tableSize{}
\centering
\begin{tabular}{Hlrgr}
\toprule
Model & Loss function & Per clip & Top 1 & Top 5 \\
\midrule
R(2+1)D-$18$ & cross-entropy (over softmax) & 37.8 & 40.3 & 74.4 \\  
R(2+1)D-$18$ & mean squared error (over logits) & 35.6 & 39.9 & 70.5 \\  
\bottomrule
\end{tabular}
\caption{
{\bf Loss function for distillation.}
We compare different loss functions for distillation, and find that the performance was relatively stable with different choices. The model used here is a 18-layer R(2+1)D, over 8-frame input, evaluated using percentage accuracy on HMDB-51 split 1.
}\label{tab:loss_fn}
\end{table}

\begin{figure}[t]
  \centering
  \begin{minipage}[c]{0.45\linewidth}
    \includegraphics[width=\linewidth]{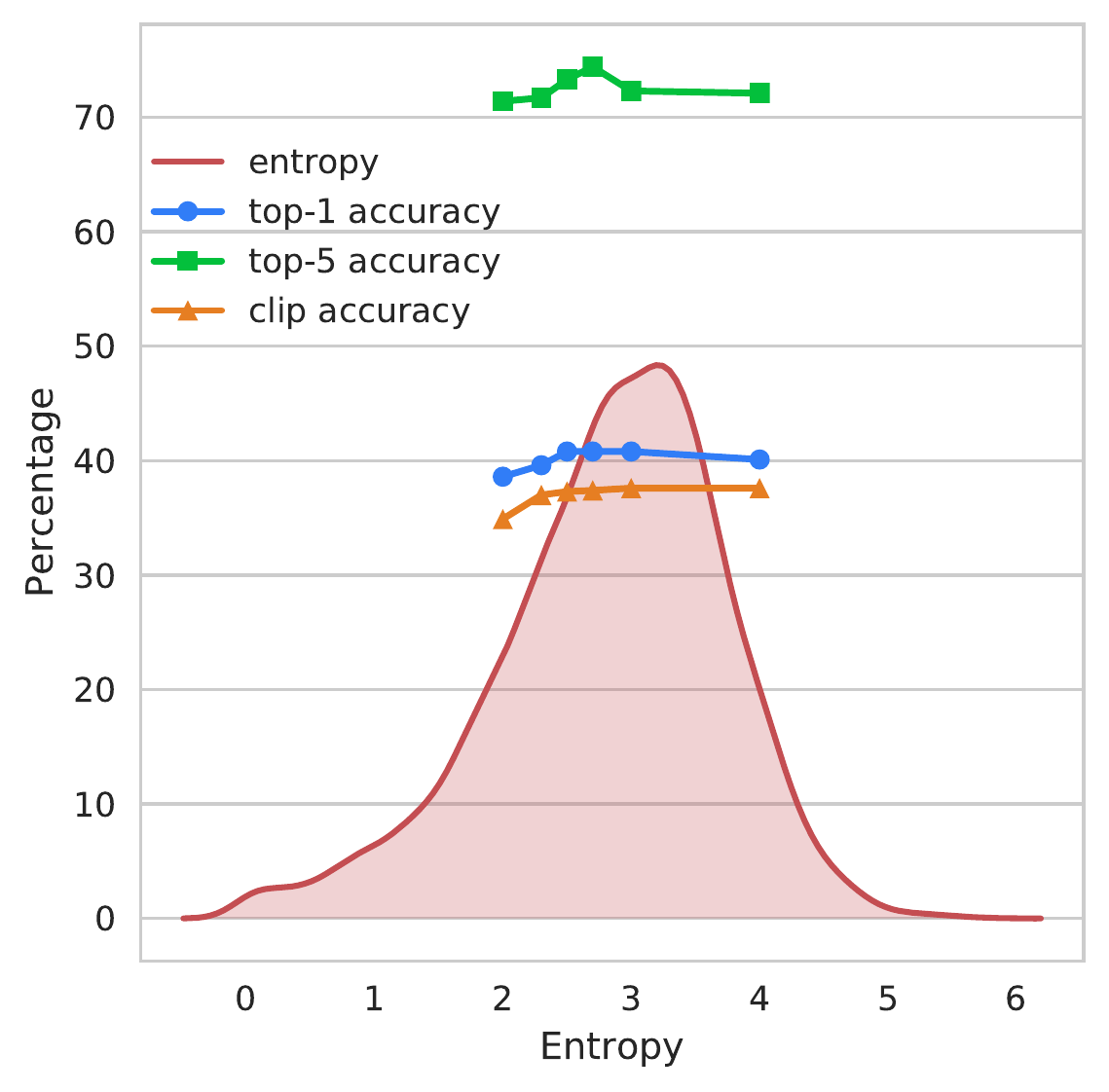}
  \end{minipage}\hfill
  \begin{minipage}[c]{0.54\linewidth}
    \caption{
    {\bf Accuracy variation with entropy.}
    Our results suggest that if we use \METHOD{} only on videos for which the teacher is sufficiently sure of its predictions (entropy $< 2.7$), we obtain slightly better performance while ignoring 50\% of the input videos, making training faster.
    }\label{fig:entropy}
  \end{minipage}
\end{figure}

{\noindent \bf Selecting confident predictions:}
As some recent work~\cite{radosavovic2018data} has shown, distillation techniques can benefit from using only
the most confident predictions for training the student. We use the entropy of the predictions from the teachers as
a notion of their confidence. 
We implement this confidence thresholding by setting a zero weight for the loss on each example, for which the 
teacher is not confident, or has high-entropy predictions; effectively dropping parts of the training data
that are confusing for the teacher.
We show the performance on dropping different amounts of data in Figure~\ref{fig:entropy}.
The red curve shows a kernel density estimate (a PDF) of entropy values for an ImageNet teacher on the Kinetics data. At any given entropy value ($e$), it shows the relative likelihood of a data point to have that entropy value, and $\int_{-\infty}^{e} f(x)dx$ (area under the curve from $-\infty$ to $e$) is the percentage of data with entropy $\leq e$. We experiment with setting different thresholds for dropping the low-confidence data points during \METHOD{}, and show the downstream HMDB-51 split-1 performance in the line plots.
We found slightly better performance, even after dropping nearly half the data, making training faster.

{\noindent \bf Varying the unlabeled dataset:}\label{sec:expt:distill_data}
We now try to evaluate whether our method is dependent on any specific video data source, and if it can benefit from additional
data sources. We evaluate this in Table~\ref{tab:data} and observe nearly similar performance when using
different sets of videos (without labels) like Kinetics~\cite{kay2017kinetics} and Sports1M~\cite{Karpathy_14}.
We also experiment with an internal set of videos downloaded from the web, and still get strong \METHOD{} performance. This shows our method is not limited to any 
form of data curation, and can learn from truly in-the-wild videos.

\begin{table}[t]
\setlength\tabcolsep{6pt} 
\tableSize{}
\centering
\begin{tabular}{lllrgr}
\toprule
Model & Unlabeled set & Size & Per clip & Top 1 & Top 5 \\
\midrule
R(2+1)D-$18$ & Kinetics~\cite{kay2017kinetics} & 0.3M & 37.8 & 40.3 & 74.4 \\  
R(2+1)D-$18$ & Sports1M~\cite{Karpathy_14} & 1.1M & 37.5 & 39.9 & 73.3 \\  
R(2+1)D-$18$ & Kinetics+Sports1M & 1.4M & 38.0 & 41.8 & 75.3 \\  
R(2+1)D-$18$ & Internal videos & 1.0M & 38.2 & 41.2 & 72.0 \\
\bottomrule
\end{tabular}
\caption{
{\bf Unlabeled data for distillation.}
This table shows that our model is not limited to any specific source of unlabeled data, and can also benefit from multiple sources of data. Size denotes the number of unlabeled videos used from that set. Performance reported on HMDB-51 split 1.
}\label{tab:data}
\end{table}

{\noindent \bf Using other teachers:}
Just as our model is capable of learning from more data, our model is also capable of using 
diverse supervision.
We experiment with replacing the ImageNet teacher with a model trained on PlaceNet~\cite{zhou2017places}, and obtain 36.8\%
HMDB fine-tuning performance as opposed to 40.3\% before with ImageNet. 
Apart from the fact that our model can learn from diverse sources of supervision, this result shows
that objects $\rightarrow$ actions semantic transfer is more effective than scenes $\rightarrow$ actions.
This makes sense as human actions are typically informed more by the objects in their environment, than the environment itself.
We also tried training with both ImageNet and PlaceNet teachers
jointly, and obtained a top-1 accuracy of 40.7\%, suggesting
that there is little benefit of adding scene
cues (from Places) given object information (from ImageNet). However, teachers from unrelated domains are likely
to provide more complementary information and lead to higher improvements.

\begin{table}[t]
\setlength\tabcolsep{6pt} 
\tableSize{}
\centering
\begin{tabular}{llrgr}
\toprule
Model & Initialization & Per clip & Top 1 & Top 5 \\
\midrule
Res3D-$18$ & Scratch & 30.7 & 38.7 & 70.0 \\  
Res3D-$18$ & ImageNet mean inflated & 33.5 & 43.9 & 73.9 \\  
R(2+1)D-$18$ & DistInit (ours) & {\bf 42.6} & {\bf 49.2} & {\bf 81.2} \\  
\arrayrulecolor{black}
\bottomrule
\end{tabular}
\caption{
{\bf \METHOD{} on Optical Flow.}
This table shows that our model is also applicable to other modalities, like optical flow.
Note that the inflated initialization for the first layer ({\tt conv\_1}) was performed by averaging
the kernel on channel dimension, and then replicating it two times. 
Reported on HMDB-51 split 1.
}\label{tab:flow}
\end{table}

\begin{table}[t]
\setlength\tabcolsep{6pt} 
\tableSize{}
\centering
\resizebox{\linewidth}{!}{
\begin{tabular}{lllllrr}
\toprule
Model & Architecture & \#frames & Pre-training & Split 1 & 3-split avg \\
\midrule
Misra {\it et al.}~\cite{misra2016unsupervised} & AlexNet~\cite{krizhevsky2012imagenet} & 1 & Scratch & - & 13.3 \\
Misra {\it et al.}~\cite{misra2016unsupervised} & AlexNet~\cite{krizhevsky2012imagenet} & 1 & Tuple verify~\cite{misra2016unsupervised} & - & 18.1 \\
Misra {\it et al.}~\cite{misra2016unsupervised} & AlexNet~\cite{krizhevsky2012imagenet} & 1 & ImageNet & - & 28.5 \\
Two-stream (RGB)~\cite{Simonyan_14b,twofusion_web} & VGG-M~\cite{Simonyan_14a} & 1 & ImageNet & - & 40.5 \\
C3D~\cite{carreira2017quo} & Custom & 16 & Scratch & 24.3 & - \\
\arrayrulecolor{GrayLine}
\midrule
LSTM~\cite{carreira2017quo} & BN-Inception~\cite{Ioffe_15} & - & ImageNet & 36.0 & - \\
Two stream (RGB)~\cite{carreira2017quo} & BN-Inception~\cite{Ioffe_15} & 1 & ImageNet & 43.2 & - \\
I3D (RGB)~\cite{carreira2017quo} & BN-Inception~\cite{Ioffe_15} & 64 & ImageNet & 49.8 & - \\
\midrule
Ours (RGB) & R(2+1)D-18~\cite{tran2018closer} & 32 & \METHOD{} & {\bf 54.9} & {\bf 54.8} \\  
\arrayrulecolor{black}
\bottomrule
\end{tabular}
}
{(a) HMDB-51}

\resizebox{\linewidth}{!}{
\begin{tabular}{lllllrr}
\toprule
Model & Architecture & \#frames & Pre-training & Split 1 & 3-split avg \\
\midrule
Misra {\it et al.}~\cite{misra2016unsupervised} & AlexNet~\cite{krizhevsky2012imagenet} & 1 & Scratch & - & 38.6 \\
Misra {\it et al.}~\cite{misra2016unsupervised} & AlexNet~\cite{krizhevsky2012imagenet} & 1 & Tuple verification~\cite{misra2016unsupervised} & - & 50.2 \\
Two-stream (RGB)~\cite{Simonyan_14b,twofusion_web} & VGG-M~\cite{Simonyan_14a} & 1 & ImageNet & - & 73.0 \\
C3D~\cite{carreira2017quo} & Custom & 16 & Scratch & 51.6 & - \\
\arrayrulecolor{GrayLine}
\midrule
LSTM~\cite{carreira2017quo} & BN-Inception~\cite{Ioffe_15} & - & ImageNet & 81.0 & - \\
Two stream (RGB)~\cite{carreira2017quo} & BN-Inception~\cite{Ioffe_15} & 1 & ImageNet & 83.6 & - \\
I3D (RGB)~\cite{carreira2017quo} & BN-Inception~\cite{Ioffe_15} & 64 & ImageNet & 84.5 & - \\
STC~\cite{diba2018spatio} & 3D-ResNet & 16 & Knowledge Tx & 82.1 & - \\
STC~\cite{diba2018spatio} & STC-ResNext~\cite{diba2018spatio} & 16 & Knowledge Tx & 84.7 & - \\
\midrule
Ours (RGB) & R(2+1)D-18~\cite{tran2018closer} & 32 & \METHOD{} & {\bf 85.7} & {\bf 85.8} \\  
\arrayrulecolor{black}
\bottomrule
\end{tabular}
}
{(b) UCF-101}
\caption{
{\bf Comparison with previous work on HMDB and UCF.}
We split the tables based on the base architecture for fair comparison.
In the first section, we report architectures with comparable depth as ours,
and in the second we report other approaches using deeper architectures.
Our model
out-performs all these previous methods.
Note that we do not compare to Kinetics pre-trained models.
Using Kinetics for pre-training with I3D~\cite{carreira2017quo} gets 74.3\% and 95.1\% 3-split avg on HMDB and UCF, but is not comparable to our unsupervised approach which does not use those labels.
}\label{tab:sota-ucf}
\end{table} 
{\noindent \bf Different input modalities:}\label{sec:expt:flow}
One of the biggest advantages of our method is that it is applicable to learn representations for any arbitrary
input data modality. We experiment with optical flow, which still contributes to significant performance
improvements on video tasks, even with modern
video architectures, across different datasets~\cite{carreira2017quo}. 
Previous work~\cite{carreira2017quo,WangL_16a} has used ImageNet initialization for networks accepting flow as input.
This is far from ideal since flow has much different statistics than RGB images. \METHOD{},
on the other hand, is agnostic to the input data modality of the student network. We train the student network
to learn from the input flow modality, while the teacher uses a random RGB frame from the same clip
to generate the distillation target.
As we show in Table~\ref{tab:flow},
\METHOD{} still produces strong initialization and improves over training from scratch or the ImageNet inflated
initialization. However, due to high computational cost of computing flow, we ignore this input modality for the final comparisons.

\subsection{Comparison with previous work}\label{sec:expt:sota}
\begin{figure}
    \centering
    \includegraphics[width=\linewidth]{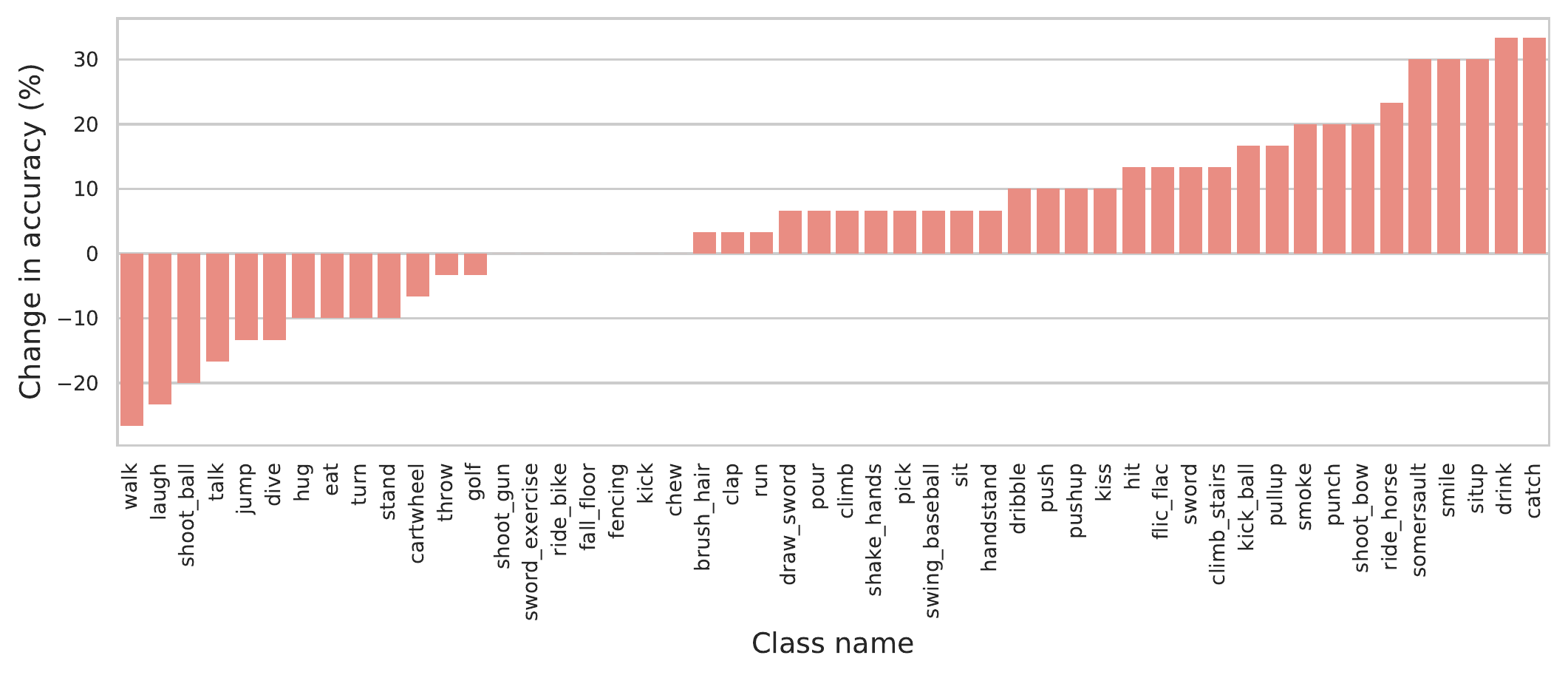}
    \caption{{\bf HMDB classes with largest gain using \METHOD{} instead of inflation.}
        The plot shows HMDB per-class accuracy difference between the two finetuned models tested. It suggests that our method is most useful for classes that require understanding motion, such as ``catching'' and  ``smiling.'' On the other hand, classes like ``shoot\_ball'' or ``eat'' are easy to recognize from single frames.
    }
    \label{fig:improve}
\end{figure}

Finally, in Table~\ref{tab:sota-ucf} we compare our model to other standard models and initialization methods on 
HMDB and UCF.
For these comparisons, we use the
32-frames model, tested using dense predictions (instead
of uniformly sampled 10-clips) for each testing video.
Here we only compare to RGB-based models for computational speed, though our approach is applicable 
to flow or other modalities as shown in Section~\ref{sec:expt:flow}.
We obtain strong performance compared to standard methods, and other unsupervised
feature learning techniques~\cite{misra2016unsupervised,diba2018spatio}.
Finally, in Figure~\ref{fig:improve} we show which classes benefit the most from the initialization provided by \METHOD{} compared to that computed by inflation.

 \section{Conclusion}\label{sec:concl}

We describe a simple approach to transfer knowledge from image-based datasets labeled for object or scene recognition tasks, to learn spatiotemporal video models for human action recognition tasks. Much previous work has addressed this problem by constraining spatiotemporal architectures to match  2D counterparts, limiting the choice of networks that can be explored.  We describe a simple approach, \METHOD{}, based on distillation that can be used to initialize any spatiotemporal architecture. It does so by making use of image-based teachers that can leverage considerable knowledge about objects, scenes, and potentially other semantics (e.g., attributes, pose) encoded in richly-annotated image datasets.
Unlike previous unsupervised learning works that depend on the curated ImageNet dataset, albeit without labels, we show our model even works on truly in-the-wild uncurated videos.
We demonstrate significant improvements over standard best practices for initializing spatiotemporal models. That said, our results do not match the accuracy of models pretrained on recently-introduced, large-scale {\em supervised} video datasets. But we note that these were collected and annotated with significant manual effort. Because our approach requires only {\em unsupervised} videos, it has the potential to make use of massively-larger data for learning accurate video models.

 {\small
\paragraph{Acknowledgments:}
RG and DR were partly supported by the Intelligence Advanced Research Projects Activity (IARPA) via Department of Interior/Interior Business Center (DOI/IBC) contract number D17PC00345. The U.S. Government is authorized to reproduce and distribute reprints for Governmental purposes not withstanding any copyright annotation theron. Disclaimer: The views and conclusions contained herein are those of the authors and should not be interpreted as necessarily representing the official policies or endorsements, either expressed or implied of IARPA, DOI/IBC or the U.S. Government.
} 
{\small
\bibliographystyle{ieee_fullname}
\bibliography{refs}
}

\end{document}